\documentclass[runningheads, a4paper]{llncs}
\usepackage[misc]{ifsym}
\usepackage[latin9]{inputenc}
\PassOptionsToPackage{vlined, ruled}{algorithm2e}
\usepackage{color}
\usepackage{float}
\usepackage{textcomp}
\usepackage{algorithm2e}
\usepackage{amsmath}
\usepackage[pdftex]{graphicx}
\usepackage{epstopdf}
\PassOptionsToPackage{normalem}{ulem}
\usepackage{ulem}
\usepackage[unicode=true,pdfusetitle,
 bookmarks=true,bookmarksnumbered=false,bookmarksopen=false,
 breaklinks=false,pdfborder={0 0 1},backref=false,colorlinks=true]
 {hyperref}

\makeatletter

\pdfpageheight\paperheight
\pdfpagewidth\paperwidth

\newcommand{\lyxmathsym}[1]{\ifmmode\begingroup\def\b@ld{bold}
  \text{\ifx\math@version\b@ld\bfseries\fi#1}\endgroup\else#1\fi}

\providecommand{\tabularnewline}{\\}
\newcommand{\lyxdot}{.}

\usepackage{amsmath}
\usepackage{amssymb}
\usepackage{amsfonts}
\DeclareMathOperator*{\argmax}{argmax}

\LinesNumbered

\@ifundefined{showcaptionsetup}{}{%
 \PassOptionsToPackage{caption=false}{subfig}}
\usepackage{subfig}
\makeatother

\begin{document}
\title{Bayesian Optimization with Missing Inputs \thanks{To appear at ECML-PKDD 2020 conference.}}
\titlerunning{Bayesian Optimization with Missing Inputs}

\toctitle{Bayesian Optimization with Missing Inputs}

\author{Phuc Luong \Letter \and Dang Nguyen \and Sunil Gupta \and Santu Rana \and \\Svetha Venkatesh}
\authorrunning{P. Luong et al.}

\tocauthor{Phuc Luong (Deakin University),
Dang Nguyen (Deakin University),
Sunil Gupta (Deakin University),
Santu Rana (Deakin University),
Svetha Venkatesh (Deakin University)}

\institute{\email{$\{$pluong, d.nguyen, sunil.gupta, santu.rana, svetha.venkatesh$\}$@deakin.edu.au}\\
Applied Artificial Intelligence Institute $\left(\textrm{A}^{2}\textrm{I}^{2}\right)$\\
Deakin University, Waurn Ponds, Geelong, VIC 3216, Australia}
\maketitle \setcounter{footnote}{0}
\begin{abstract}
\label{sec:abstract} Bayesian optimization (BO) is an efficient method for optimizing expensive
black-box functions. In real-world applications, BO often faces a
major problem of missing values in inputs. The missing inputs can
happen in two cases. First, the historical data for training BO often
contain missing values. Second, when performing the function evaluation
(e.g. computing alloy strength in a heat treatment process), errors
may occur (e.g. a thermostat stops working) leading to an erroneous
situation where the function is computed at a random unknown value
instead of the suggested value. To deal with this problem, a common
approach just simply skips data points where missing values happen.
Clearly, this naive method cannot utilize data efficiently and often
leads to poor performance. In this paper, we propose a novel BO method
to handle missing inputs. We first find a probability distribution
of each missing value so that we can impute the missing value by drawing
a sample from its distribution. We then develop a new acquisition
function based on the well-known Upper Confidence Bound (UCB) acquisition
function, which considers the uncertainty of imputed values when suggesting
the next point for function evaluation. We conduct comprehensive experiments
on both synthetic and real-world applications to show the usefulness
of our method.

\end{abstract}

\keywords{Bayesian optimization \and Missing data \and Matrix
factorization \and Gaussian process.}

\section{Introduction}

\label{sec:intro} Bayesian optimization (BO) \cite{shahriari2016taking} is a powerful
tool to optimize expensive black-box functions. Typically, at each
iteration BO first models the black-box function via a statistical
model e.g. a Gaussian process (GP) based on historical data (\textit{observed
data}) and then seeks out the next point (\textit{suggestion}) for
function evaluation by maximizing an \textit{acquisition function}.
BO has been successfully applied to a wide range of practical applications
such as hyper-parameter tuning, automated machine learning, material
design, and robot exploration \cite{snoek2012practical,nguyen2020bayesian,frazier2016bayesian,oliveira2019bayesian}.

\begin{figure}[H]
\begin{centering}
\includegraphics[scale=0.32]{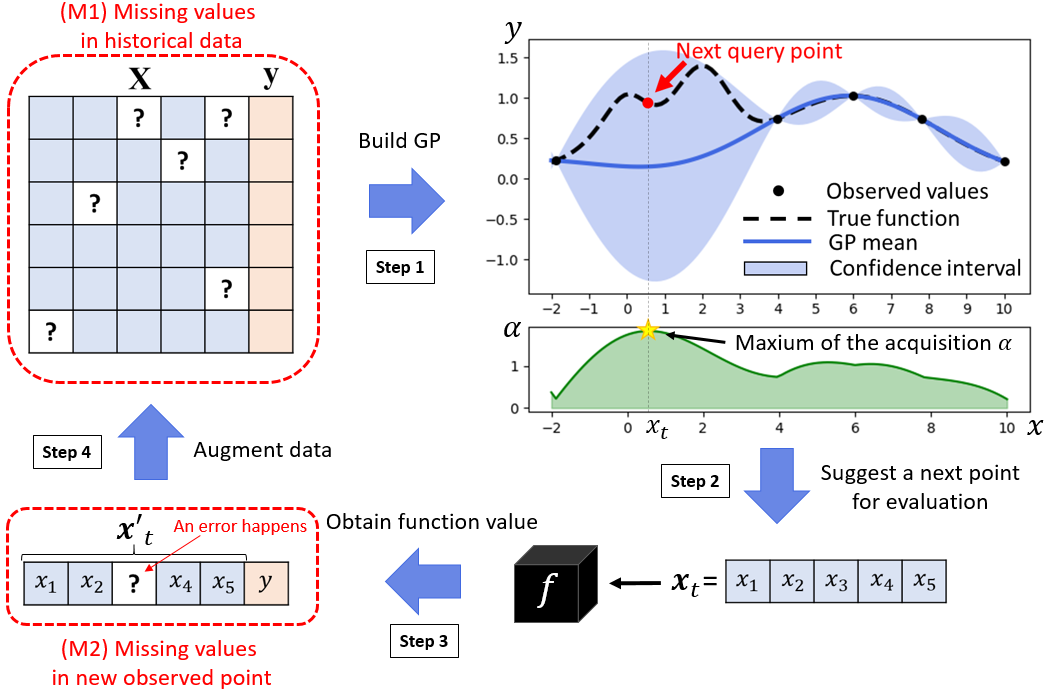}
\par\end{centering}
\caption{\label{fig:BO-in-optimization-with-missing}Four main steps in BO:
(1) build a GP from the historical data $\left[\mathbf{X},\mathbf{y}\right]$,
(2) maximize an acquisition function $\alpha$ to get a suggested
point $\boldsymbol{x}_{t}$, (3) evaluate the suggested point $\boldsymbol{x}_{t}$
with the true black-box function and obtain a function value $y$,
and (4) augment the historical data with the new observed point. With
the presence of missing values in (\textbf{M1})\textit{ historical
data} and (\textbf{M2})\textit{ new observed point}, BO faces two
significant problems: it cannot build the GP at Step-1 and it cannot
use the new observed point at Step-4.}

\end{figure}

In real-world applications, BO often faces a significant problem that
is \textit{missing values in inputs}. As shown in Figure \ref{fig:BO-in-optimization-with-missing},
missing values in input can happen in two cases. First, similar to
other machine learning models, the historical data for training BO
may contain missing values (\textit{missing values in historical data}).
Without imputing these missing values, we cannot model the black-box
function using GP. Second, when performing the function evaluation
at the suggested point, if an error happens (e.g. failure of devices),
we obtain the function value at an unknown random point (\textit{missing
values in new observed point}). Missing values
in input can lead to many crucial failures in BO optimization such
as erroneous calculation, and difficulties in interpretation and representation
of information \cite{smieja2019generalized}.

To address the missing input problem in BO, one approach is to apply
imputation methods e.g. mean/mode imputation and k-nearest neighbors
(KNN) \cite{ambler2007comparison,bertsimas2017predictive} to fill
missing values first and then apply a traditional BO method. Although
these imputation methods can predict missing values, their performance
is non-optimal since mean/mode methods do not consider the correlation
between missing values and non-missing values while KNN strongly depends
on the current available data and distance metric \cite{donders2006gentle}.
Recently, more complex imputation methods have been introduced, including
using random forest \cite{tang2017random} and deep neural network
\cite{yoon2018gain}; however, these methods require extensive training
data, which is unrealistic in the BO context where the historical
data is quite limited. Another approach is to simply apply BO to non-missing
data where points with missing values are removed \cite{kang2013prevention}.
As mentioned before, this approach does not use the data efficiently,
leading to poor performance in optimization. Oliveira et. al proposed
a BO method for uncertain inputs \cite{oliveira2019bayesian}, where
they observe the difference between the actual input value and the
one recommended by BO, and they estimate the variance needed to build
the probability distribution of input values. However, in the case
of missing values, this variance is unknown (i.e. the noise level
added to the actual input value is unknown), their method cannot approximate
missing values well. To the best of our knowledge, there is no BO
method that can directly handle missing values in input.

\textbf{Our method.} To overcome the disadvantages of existing methods,
we propose a novel method (named \textbf{BOMI}) for optimizing black-box
functions with missing values in input. In particular, we first adapt
the idea of Bayesian probabilistic matrix factorization (BPMF) \cite{salakhutdinov2008bayesian}
to find the distribution of each missing value for imputation. Note
that none of the imputation methods discussed above use the distributions
of missing values for imputation even though these distributions are
essential since they represent a certain level of noise in the actual
values. By adapting the idea of BPMF, these distributions are built
using one of the collaborative filtering technique so that the correlation
between values in the data is taken into account. We then propose
a new acquisition function, based on the widely used UCB acquisition
function \cite{srinivas2012information}, to achieve greater confidence
in modeling the black-box function. Our new acquisition function differs
from a traditional acquisition function in a sense that it does not
use one single GP built from imputed data but leverages multiple GPs
to take into account the uncertainty of predicted values. By doing
this, our method achieves an agreement on the imputed values that
results in a higher confidence in the posterior predictive distribution.
As a result, it improves the optimization performance when the black-box
functions involve missing inputs.

To summarize, we make the following contributions.
\begin{itemize}
\item Develop \textit{\uline{B}}\textit{ayesian }\textit{\uline{O}}\textit{ptimization
with }\textit{\uline{M}}\textit{issing }\textit{\uline{I}}\textit{nputs}
(\textbf{BOMI}) to optimize black-box functions with missing values
in input. 
\item Propose a new acquisition function that takes into account the distributions
of missing values when suggesting the next point for function evaluation.
\item Demonstrate the usefulness of \textbf{BOMI} in both synthetic and
real-world applications, and show that it outperforms well-known state-of-the-art
baselines.
\end{itemize}

\section{Background}

\label{sec:bg} 
\subsection{Bayesian optimization}

Bayesian optimization (BO) is an efficient method for automatically
finding the optimum of an expensive black-box function within a small
number of function evaluations \cite{mockus1978application,brochu2010tutorial}.
Given an unknown function $f:\mathcal{X}\rightarrow\mathbb{R}$, our
goal is to find the optimal input $x^{*}=\argmax_{x\in\mathcal{X}}\,f\left(x\right)$,
where $\mathcal{X}$ is a bounded domain in $\mathbb{R}^{d}$. Since
the objective function $f$ is expensive to evaluate, BO attempts
to model $f$ via a surrogate model e.g. Gaussian process (GP) \cite{rasmussen2003gaussian}.
The function $f$ is assumed to be drawn from the GP, i.e. $f(x)\sim\mathcal{GP}(\mu(x),k(x,x'))$,
where $\mu:\mathcal{X}\rightarrow\mathbb{R}$ and $k:\mathcal{X}\mathcal{\times X}\rightarrow\mathbb{R}$
are mean and covariance functions. Normally, $\mu\left(x\right)$
is assumed to be zero and $k$ is the \textit{squared exponential}
kernel (Equation (\ref{eq:ise})):
\begin{equation}
k(x,x')=\sigma^{2}exp(-\frac{1}{2l^{2}}\|x-x'\|^{2})\label{eq:ise}
\end{equation}
where $\sigma^{2}$ is a parameter dictating the uncertainty in $f$,
and $l$ is a length scale parameter which controls how quickly $f$
can change.

Given the historical data up to iteration $t$, $\mathcal{D}_{t}=\{(x_{i},y_{i})\}_{i=1}^{t}$
that contains inputs $x_{i}$ and their evaluations $y_{i}=f(x_{i})+\epsilon_{i}$
for $i=1,2,\ldots,t$ where $\epsilon_{i}\sim\mathcal{N}(0,\sigma_{\epsilon}^{2})$,
we obtain the predictive distribution $f\left(x\right)\mid\mathcal{D}_{t}\sim\mathcal{N}\left(\mu_{t}(x),\sigma_{t}^{2}\left(x\right)\right)$
with $\mu_{t}(x)$ and $\sigma_{t}^{2}\left(x\right)$ as:
\begin{equation}
\mu_{t}(x)=\textbf{k}^{T}(\textbf{K}+\sigma_{\epsilon}^{2}\textbf{I})^{-1}\textbf{y}\label{eq:meanGP}
\end{equation}
\begin{equation}
\sigma_{t}^{2}(x)=k(x,x)-\textbf{k}^{T}(\textbf{K}+\sigma_{\epsilon}^{2}\textbf{I})^{-1}\textbf{k}\label{eq:covarGP}
\end{equation}
where $\textbf{y}=(y_{1},...,y_{t})$ is a vector of function evaluations,
$\textbf{k}=[k(x_{i},x)]_{\forall x_{i}\in\mathcal{D}_{t}}$ is the
covariance between a new input $x$ and all observed inputs $x_{i}$,
$\textbf{K}=[k(x_{i},x_{j})_{\forall x_{i},x_{j}\in\mathcal{D}_{t}}]$
is the covariance matrix between all inputs, \textbf{I} is an identity
matrix with the same dimension as \textbf{K}, and $\sigma_{\epsilon}^{2}$
is a measurement noise.

BO uses the predictive mean and standard deviation in Equations (\ref{eq:meanGP})
and (\ref{eq:covarGP}) in an \textit{acquisition function} $\alpha(x)$
to find the next point to evaluate. The acquisition function uses
the predictive distribution to balance two contrasting goals: sampling
where the function is expected to have a high value vs. sampling where
the uncertainty about the function value is high. Some well-known
acquisition functions are \textit{Probability of Improvement} (PI)
\cite{kushner1964new}, \textit{Expected Improvement} (EI) \cite{jones1998efficient},
\textit{Upper Confidence Bound} (UCB) \cite{srinivas2012information},
and \textit{Predictive Entropy Search} (PES) \cite{hernandez2014predictive}.

Since we use UCB as a base to develop a new acquisition function (see
Section \ref{subsec:BOMI}) for the optimization problem with missing
inputs, we describe it in detail in the next section.

\subsection{Upper Confidence Bound acquisition function}

The UCB acquisition function is a weighted sum of predictive mean
and variance from Equations (\ref{eq:meanGP}) and (\ref{eq:covarGP}),
computed as:
\begin{equation}
\alpha_{t}^{UCB}(x)=\mu_{t}(x)+\sqrt{\beta_{t}}\sigma_{t}(x)\label{eq:gpucb}
\end{equation}
where $\beta_{t}$ is the exploitation-exploration trade-off factor.
Following \cite{srinivas2012information}, $\beta_{t}$ is calculated
as $\beta_{t}=2\log\left(t^{2}2\pi^{2}/3\delta\right)+2d\log\left(t^{2}dbr\sqrt{\log\left(4da/\delta\right)}\right)$
to guarantee an upper bound on the cumulative regret with probability
greater than $1-\delta$ in the search space $\mathcal{X}\subseteq\left[0,r\right]^{d}$,
where $r>0$ and $a,b>0$ are constants.

To suggest a next point for the black-box objective function evaluation,
we maximize the UCB acquisition function in Equation (\ref{eq:gpucb})
as follows:
\begin{equation}
x_{t+1}=\argmax_{x\in\mathcal{X}}\,\alpha_{t}^{UCB}(x)\label{eq:maximizeacq}
\end{equation}

\section{Framework}

\label{sec:framework} 
\subsection{Problem definition}

Before formally defining the problem of Bayesian optimization (BO)
with missing inputs, we provide two cases when missing values occur
in inputs.
\begin{case}
\label{def:1_Missing-input-observation}(\textbf{Missing values in
historical data}) Given a point $\boldsymbol{x}=$$\left\{ x_{1},\ldots,x_{d}\right\} $
in historical data, $\boldsymbol{x}$ contains missing values if $\exists x_{i}\in\boldsymbol{x}$
($i\in\{1,...,d\}$ and $d$ is the input dimension) such that $x_{i}$
is unobserved (i.e. missing), and we denote $x_{i}$ by a question
mark `?'.
\end{case}

\begin{case}
\label{def:2_Missing-during-evaluation}(\textbf{Missing values in
the next suggested point}) At iteration $t$, when we intend to evaluate
the black-box function $f$ at a suggested point $\boldsymbol{x}_{t}=$$\left\{ x_{1},\ldots,x_{d}\right\} _{t}$
to obtain the function value $y_{t}$, two scenarios may arise. (1)
Due to an error in the evaluation, the function may actually be evaluated
at $\boldsymbol{x}_{t}'$ instead of intended point $\boldsymbol{x}_{t}$.
In general, we denote an element $x_{i}'$ of $\boldsymbol{x}_{t}'$
using the corresponding element of $\boldsymbol{x}_{t}$ as $x_{i}'=x_{i}\pm\eta$
where $\eta$ is an unknown noise amount. (2) In case of \emph{no
error}, $\boldsymbol{x}_{t}'$ is same as $\boldsymbol{x}_{t}$.
\end{case}

We present the problem of BO with missing inputs. Given a historical
data $\left[\mathbf{X},\mathbf{y}\right]$ and a \textit{black-box}
function $f:\mathcal{X}\rightarrow\mathbb{R}$ ($\mathcal{X}$ is
the input domain), $\mathbf{X}$ may contain missing values as mentioned
in Case \ref{def:1_Missing-input-observation}, and if we query a
point \textbf{$\boldsymbol{x}_{t}\in\mathcal{X}$} to compute the
function value $y_{t}=f\left(\boldsymbol{x}_{t}\right)$, then we
may obtain $y_{t}=f\left(\boldsymbol{x}_{t}'\right)$ as mentioned
in Case \ref{def:2_Missing-during-evaluation}. Our goal is to find
the optimal point $\boldsymbol{x}^{*}$ that maximizes the black-box
function $f$, as follows:
\begin{equation}
\boldsymbol{x}^{*}=\arg\max_{x\in\mathcal{X}}f(\boldsymbol{x})\label{eq:def_problem_func_with_missing-1}
\end{equation}

\subsection{Building a probability distribution for each missing value\label{subsec:Probability-missing}}

Let $\boldsymbol{x}_{o}$ and $\boldsymbol{x}_{m}$ be non-missing
and missing values. An observation is denoted as $\boldsymbol{x}=\left\{ \boldsymbol{x}_{o},\boldsymbol{x}_{m}\right\} $.
To solve the optimization problem in Equation (\ref{eq:def_problem_func_with_missing-1}),
one simple approach is to omit observations having missing values
$\boldsymbol{x}_{m}$ and then apply a standard BO to observations
containing only non-missing values. As discussed in Section \ref{sec:intro},
this method may perform poorly since it has too few data points to
build a good model. To overcome this, we propose to use the distribution
of a missing value so that we can both impute it as well as utilize
the uncertainty in its prediction. Therefore, instead of directly
substituting $x=c$ ($x\in\boldsymbol{x}_{m}$ and $c$ is a single
constant value), we assume that $x\sim p\left(x\right)$, where $p\left(x\right)$
is an unknown probability distribution of $x$, and our goal is to
find $p\left(x\right)$ for each $x\in\boldsymbol{x}_{m}$.

We represent the observed data $\left[\mathbf{X},\mathbf{y}\right]$
as a matrix $R=\left[\mathbf{X},\mathbf{y}\right]\in\mathbb{R}^{N\times M+1}$,
where $N$ is the number of rows (data points) and $M$ is the number
of columns (features). Let $x_{ij}$ be a missing value at row $i$
and column $j$, and$x_{ij}$ is assumed to be sampled from a normal
distribution $p\left(x_{ij}\right)=\mathcal{N}\left(\mu_{x_{ij}},\sigma_{x_{ij}}^{2}\right)$.
To find the distribution $p\left(x_{ij}\right)$, we adapt the idea
of Bayesian probabilistic matrix factorization (BPMF) \cite{salakhutdinov2008bayesian}.

Our goal is to decompose the partially-observed matrix $R\in\mathbb{R}^{N\times M+1}$
into a product of two smaller matrices $U\in\mathbb{R}^{N\times K}$
and $V\in\mathbb{R}^{K\times M+1}$ such that $R\approx UV$ i.e.
we find two matrices $U$ and $V$ whose product is as close as possible
to the original matrix $R$.

We first construct the prior distributions on $U$ and $V$ as follows:
\begin{equation}
\begin{array}{c}
p\left(U\mid\mu_{U},\Lambda_{U}\right)=\prod_{i=1}^{N}\mathcal{N}\left(U_{i}\mid\mu_{U},\Lambda_{U}^{-1}\right)\\
p\left(V\mid\mu_{V},\Lambda_{V}\right)=\prod_{j=1}^{M+1}\mathcal{N}\left(V_{i}\mid\mu_{V},\Lambda_{V}^{-1}\right)
\end{array}\label{eq:bg_bpmf_priorUandV}
\end{equation}
where $\Theta_{U}=\left\{ \mu_{U},\Lambda_{U}\right\} $ and $\Theta_{V}=\left\{ \mu_{V},\Lambda_{V}\right\} $
are hyper-parameters of the priors. We learn them using Gibbs sampling
\cite{neal1993probabilistic}.

Next, we sample $U$ and $V$ from their distributions, as in Equation
(\ref{eq:samplingUV}):
\begin{equation}
\begin{gathered}U_{i}^{l+1}\sim p\left(U_{i}\mid R,V^{l},\Theta_{U}^{l}\right)\textrm{ for }i=1,\ldots,N\textrm{ rows}\\
V_{j}^{l+1}\sim p\left(V_{i}\mid R,U^{l+1},\Theta_{V}^{l}\right)\textrm{ for }j=1,\ldots,(M+1)\textrm{ columns}
\end{gathered}
\label{eq:samplingUV}
\end{equation}
where $l$ is the number of iterations used in Gibbs sampling.

Finally, we reconstruct $R\approx UV$ and the missing value $x_{ij}=R_{ij}$
is filled by a linear combination of matrix product, i.e. $x_{ij}=U_{i,:}V_{:,j}$,
where $U_{i,:}$ is the row $i$ of $U$ and $V_{:,j}$ is the column
$j$ of $V$.

Although we can impute a missing value using $x_{ij}=U_{i,:}V_{:,j}$,
using a single predicted value is not effective. Thus, we go a step
further to obtain the distribution $p\left(x_{ij}\right)$ of $x_{ij}$.
In particular, following \cite{salakhutdinov2008bayesian} we use
the Monte Carlo approximation \cite{neal1993probabilistic} to approximate
$p\left(x_{ij}\right)$ as:
\[
p\left(x_{ij}\right)\approx p\left(x_{ij}\mid U_{i,:}V_{:,j},\xi\right),
\]
where $\xi=\sigma_{R_{ij}}^{2}$ (called \textit{precision factor})
is the ``width'' of distribution covering the actual value of $x_{ij}$.
To fill/predict a missing value $x_{ij}$, we simply draw a sample
from its distribution $\tilde{x}_{ij}\sim p\left(x_{ij}\right)$ and
set $x_{ij}=\tilde{x}_{ij}$.

\subsection{Bayesian optimization with missing inputs (BOMI)\label{subsec:BOMI}}

In Section \ref{subsec:Probability-missing}, we find a distribution
$p\left(x_{ij}\right)$ for each missing value $x_{ij}$. To optimize
the black-box function $f(\boldsymbol{x})$, we can simply draw a
sample $\tilde{x}_{ij}\sim p\left(x_{ij}\right)$ to fill the missing
value $x_{ij}$, and then apply a standard BO to the new non-missing
data. We call this method \textit{Imputation-BPMF}. However, the performance
of this approach heavily depends on the quality of $\tilde{x}_{ij}\sim p\left(x_{ij}\right)$.
In other words, it does not consider the uncertainty of $\tilde{x}_{ij}$.

\begin{figure}[H]
\begin{centering}
\includegraphics[scale=0.3]{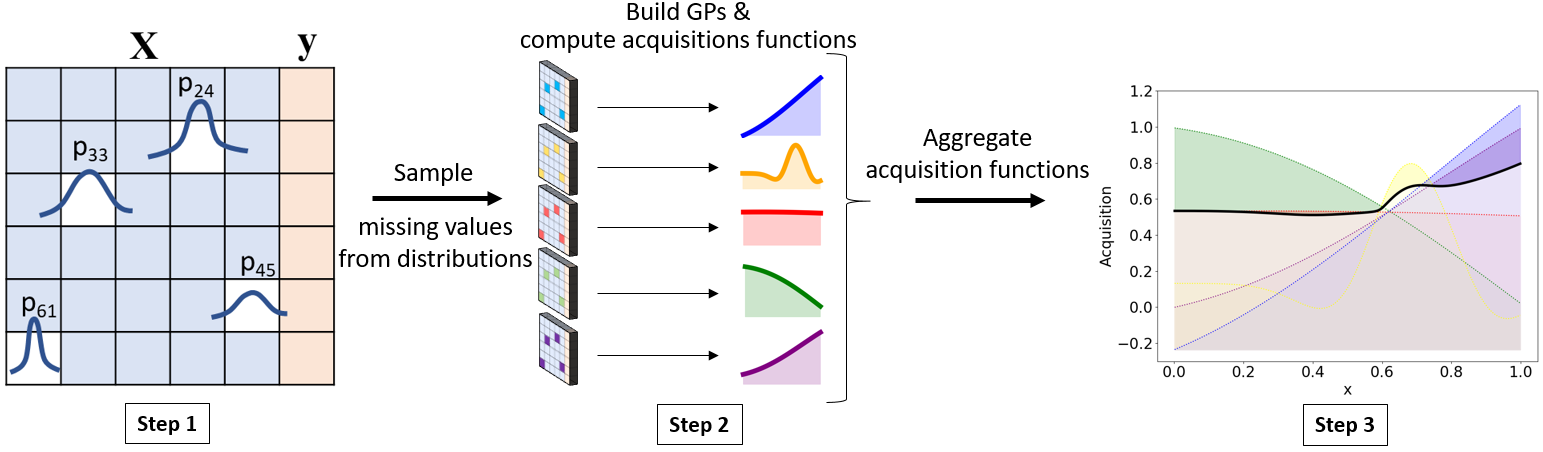}
\par\end{centering}
\caption{\label{fig:proposed_step}Three main steps in our method \textbf{BOMI}:
(1) sample missing values from their distributions, (2) build GPs
and compute UCB acquisition functions (Equation (\ref{eq:gpucb})),
and (3) develop a new acquisition function based on aggregated information.}
\end{figure}

We propose a novel BO method to optimize black-box functions with
missing inputs, called \textit{\uline{B}}\textit{ayesian }\textit{\uline{O}}\textit{ptimization
with }\textit{\uline{M}}\textit{issing }\textit{\uline{I}}\textit{nputs}
(\textbf{BOMI}). Our method has three main steps, which illustrated
in Figure \ref{fig:proposed_step}. \textbf{Step 1:} from the observed
data with missing values, \textbf{BOMI} learns a distribution for each missing
value (see Section \ref{subsec:Probability-missing}), then uses these
distributions to impute and generate $Q$ new non-missing data. \textbf{Step
2:} for each new non-missing data, \textbf{BOMI} builds a GP and computes the
acquisition function UCB (see Equation (\ref{eq:gpucb})). \textbf{Step
3:} \textbf{BOMI} aggregates the information from $Q$ acquisition functions
to come up with a new acquisition function that takes into account
the uncertainty of imputed values. The new acquisition function (called
\textbf{UCB-MI}) is described next.

\subsubsection*{\uline{U}pper \uline{C}onfidence \uline{B}ound acquisition
function for \uline{M}issing \uline{I}nputs (UCB-MI).}

Our new acquisition function UCB-MI aggregates the information from
$Q$ standard UCB acquisition functions computed at Step-2, as follows:
\begin{align}
\alpha^{UCB-MI}(x) & =\mu_{\alpha}\left(\boldsymbol{\alpha}^{UCB}(x)\right)+\beta_{\alpha}\sigma_{\alpha}\left(\boldsymbol{\alpha}^{UCB}(x)\right)\label{eq:ucb-mi}\\
 & =\frac{1}{Q}\sum_{q=1}^{Q}\left(\alpha_{q}^{UCB}(x)\right)+\beta_{\alpha}\sqrt{\frac{\sum_{q=1}^{Q}\left(\alpha_{q}^{UCB}(x)-\frac{1}{Q}\sum_{q=1}^{Q}\alpha_{q}^{UCB}(x)\right)^{2}}{Q-1}}\nonumber 
\end{align}
where $\alpha^{UCB}$ is the UCB acquisition function (see Equation
(\ref{eq:gpucb})).

Our acquisition function $\alpha^{UCB-MI}$ is based on the commonly
used UCB acquisition function, but it incorporates the posterior predictive
information from different GPs. It is described as a summation of
the mean of $Q$ acquisition values $\mu_{\alpha}\left(\alpha\right)$
and their standard deviation $\sigma_{\alpha}\left(\alpha\right)$
multiplied by a trade-off factor $\beta_{\alpha}$. This acquisition
function quantifies the level of agreement between $Q$ individual
acquisition functions to determine the confidence in predicting the
outcome of an input. As a result, we have more information about the
variance of one point $\boldsymbol{x}$ and more certainty about its
outcome. To suggest a next point for evaluation, we maximize the acquisition
function $\alpha^{UCB-MI}$:
\begin{equation}
\begin{array}{c}
x_{t+1}=\underset{x\in\mathcal{X}}{\textrm{argmax}}\:\alpha^{UCB-MI}\left(x\right)\end{array}\label{eq:proposed_acq}
\end{equation}

\subsubsection*{Discussion.}

We can see that when $Q$ is set to a small value, our acquisition
function $\alpha^{UCB-MI}$ is close to the standard acquisition function
UCB $\alpha^{UCB}$. For example, with $Q=1$, the standard deviation
$\sigma_{\alpha}\left(\boldsymbol{\alpha}^{UCB}\right)=0$ and $\alpha^{UCB-MI}(x)=\alpha_{1}^{UCB}\left(x\right)+\beta_{\alpha}0=\alpha^{UCB}\left(x\right)$.

When $Q>1$, the first term $\frac{1}{Q}\sum_{q=1}^{Q}\left(\alpha_{q}^{UCB}\left(x\right)\right)$
in Equation (\ref{eq:ucb-mi}) represents the average among different
acquisition functions, which can be considered as an agreement on
different acquisition functions. In contrast, the second term represents
the disagreement on acquisition values since it is the standard deviation
measuring how much acquisition functions differ from their mean (agreement).
The trade-off factor $\beta_{\alpha}$ is used to control the balance
between agreement and disagreement. 

Our proposed method \textbf{BOMI} is summarized in Algorithm \ref{alg:BOMI-algorithm}.

\begin{algorithm}
\RestyleAlgo{ruled}
\SetAlgoVlined
\SetKwInOut{Input}{Input} 
\Input{Observed data $D_0$, \# iterations $T$, \# new non-missing data $Q$}
\SetAlgoLined
\Begin{
\For{$t=0,...,T$}{
\For{$q=1,...,Q$}{
Sample $U_{\left(q\right)}\sim p\left(U\mid R,V,\Theta_{U}\right)$ and $V_{\left(q\right)}\sim p\left(V\mid R,U,\Theta_{V}\right)$\\
Generate new non-missing data $R_{\left(q\right)} = U_{\left(q\right)}V_{\left(q\right)}$\\
Build GP $GP_{\left(q\right)}\leftarrow R_{\left(q\right)}$\\
Compute acquisition function $\alpha_{q}^{UCB}\leftarrow$ A\textsc{cquisition}$\left(GP_{\left(q\right)}\right)$
}
Compute $\alpha^{UCB-MI}$ using Equation $\left(\ref{eq:ucb-mi}\right)$\\
Suggest a next point $x_{t+1} = \argmax_{x \in \mathcal{X}}\:\alpha^{UCB-MI}\left(x\right)$\\
Evaluate the objective function $y_{t+1}=f(x_{t+1})$\\
\If {missing event}{
$x_{t+1} \rightarrow x_{t+1}^{'}$ (see Case $\ref{def:2_Missing-during-evaluation}$)\\ 
$y_{t+1}=f(x_{t+1}^{'})$
}
Augment $D_{t+1}=\{D_t, (x_{t+1}, y_{t+1})\}$\\
}
}
\caption{\textbf{\label{alg:BOMI-algorithm}}The proposed \textbf{BOMI} algorithm.}
\end{algorithm}

\section{Experimental Results}

\label{sec:experiments} We evaluate our proposed method \textbf{BOMI} in both synthetic and
real-world applications. For synthetic experiments, we test our method
with four benchmark synthetic functions to show its optimization performance
and stability. For real-world experiments, we test the performance
of our method in two real-world applications, namely, a robot exploration
simulation and a heat treatment process. In these two applications,
missing inputs often occur since the failures of robots and thermostat
are unmanageable.

\textbf{Baselines.} We compare \textbf{BOMI} with six state-of-the-art
baselines that use different ways to deal with missing values. They
are categorized into two groups:
\begin{itemize}
\item \textbf{Imputation-based methods:} These methods first use imputation
methods to predict missing values and then simply apply a standard
BO method to optimize the black-box functions. Here, we use three
well-known imputation methods in machine learning, namely, mean, mode,
and KNN \cite{bertsimas2017predictive}. The mean method (called \textit{Imputation-Mean})
replaces a missing value by the mean of its feature column. The mode
method (called \textit{Imputation-Mode}) replaces a missing value
by the mode of its feature column. The KNN method (called \textit{Imputation-KNN})
replaces a missing value by the mean value of its $k$ nearest points.
We also compare with \textit{Imputation-BPMF}, where missing values
are imputed using the BPMF method (see Section \ref{subsec:BOMI}).
\item \textbf{BO-based methods:} Since standard BO methods cannot directly
deal with missing inputs, we consider two variants of BO. \textit{DropBO}
-- whenever a data point containing missing values occurs in historical
data or new observed point, this method simply skips that data point
and applies a standard BO method to non-missing data \cite{kang2013prevention}.
\textit{SuggestBO} -- similar to DropBO this method removes data
points containing missing values in historical data; however when
a new observed point contains missing values, instead of skipping
this new observation this method still uses it but substitutes missing
values by the values suggested by the acquisition function. We also
compare with \textit{BO-uGP} \cite{oliveira2019bayesian} -- a recent
BO method proposed for optimizing black-box functions with uncertain
inputs. This method assumes that there is no missing values but all
of them are noisy. It first maps all points into distributions and
then builds a surrogate model over the distributions of points.
\end{itemize}
\textbf{Implementation details.} We implement our method \textbf{BOMI}
and all baselines using GPyTorch \cite{gardner2018gpytorch} to accelerate
matrix multiplication operations in GP inference. For a fair comparison,
in our experiments we use the same kernel (\textit{squared exponential}
kernel) and identical initial points for all methods. For Imputation-KNN,
we use the number of neighbors $k=5$ and the Euclidean distance,
following \cite{beretta2016nearest}. For BO-uGP, we use the same
hyper-parameter setting, as mentioned in the paper. For our method
\textbf{BOMI}, we set the dimension $K$ of matrices $U$ and $V$
to 15, the precision $\xi$=0.01, the number of new non-missing data
$Q=5$, and the number of iterations in Gibbs sampling $l=40$. We
repeat each method 10 times and report the average result along with
the standard error.

\subsection{Synthetic experiments}

We test our method and baselines with four benchmark synthetic functions
where their characteristics are summarized in Table \ref{table1:syntfuncchar}.

\begin{table}[h]
\centering{}\caption{\label{table1:syntfuncchar}Characteristics of synthetic functions. }
\begin{tabular}{|l|c|l|}
\hline 
\textbf{Function} & \textbf{Dimension} & \textbf{Range}\tabularnewline
\hline 
\hline 
\textit{Eggholder} & 2 & $x_{1},x_{2}\in\left[-512,512\right]$\tabularnewline
\hline 
\textit{Schubert} & 4 & $x_{i}\in\left[-10,10\right]$ for $i=1,\ldots,4$\tabularnewline
\hline 
\textit{Alpine} & 5 & $x_{i}\in\left[-10,10\right]$ for $i=1,\ldots,5$\tabularnewline
\hline 
\textit{Schwefel} & 5 & $x_{i}\in\left[-500,500\right]$ for $i=1,\ldots,5$\tabularnewline
\hline 
\end{tabular}
\end{table}

\subsubsection{Performance comparison.}

The first experiment illustrates how our method \textbf{BOMI} outperforms
other methods in terms of optimization result.

\textbf{Experiment settings.} We initialize 30 data points (historical
data) for each function and keep them the same for all methods. To
see the effect of missing values, we allow 80\% of historical data
to have missing values. When evaluating a suggested point, there is
a probability $\rho$ (called \textit{missing rate}) that the new
observed point has missing values (i.e. an error occurs, see Case
\ref{def:2_Missing-during-evaluation}). With this probability $\rho$,
an amount of noise $\eta$ (called \textit{missing noise}) is added
to the suggested value, which is calculated as $x_{i}^{'}=x_{i}\pm\eta r_{i}$,
where $x_{i}$ is the actual value suggested by the acquisition function,
$r_{i}$ is the value range of $x_{i}$, and $x_{i}^{'}$ is a random
unknown value. In our experiments, we set $\rho=0.25$ and $\eta=0.05$
for all functions. 

\begin{figure}
\centering{}\includegraphics[scale=0.24]{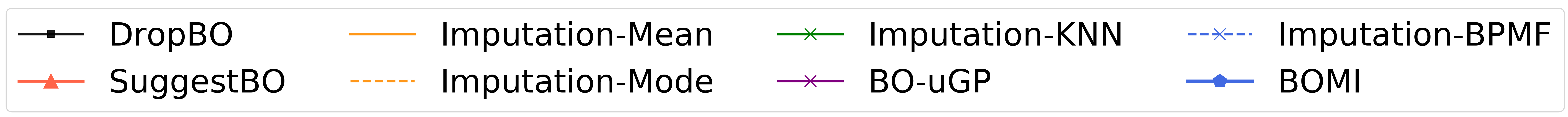}\\
\subfloat[\label{fig:performSynt_a}]{\begin{centering}
\includegraphics[scale=0.12]{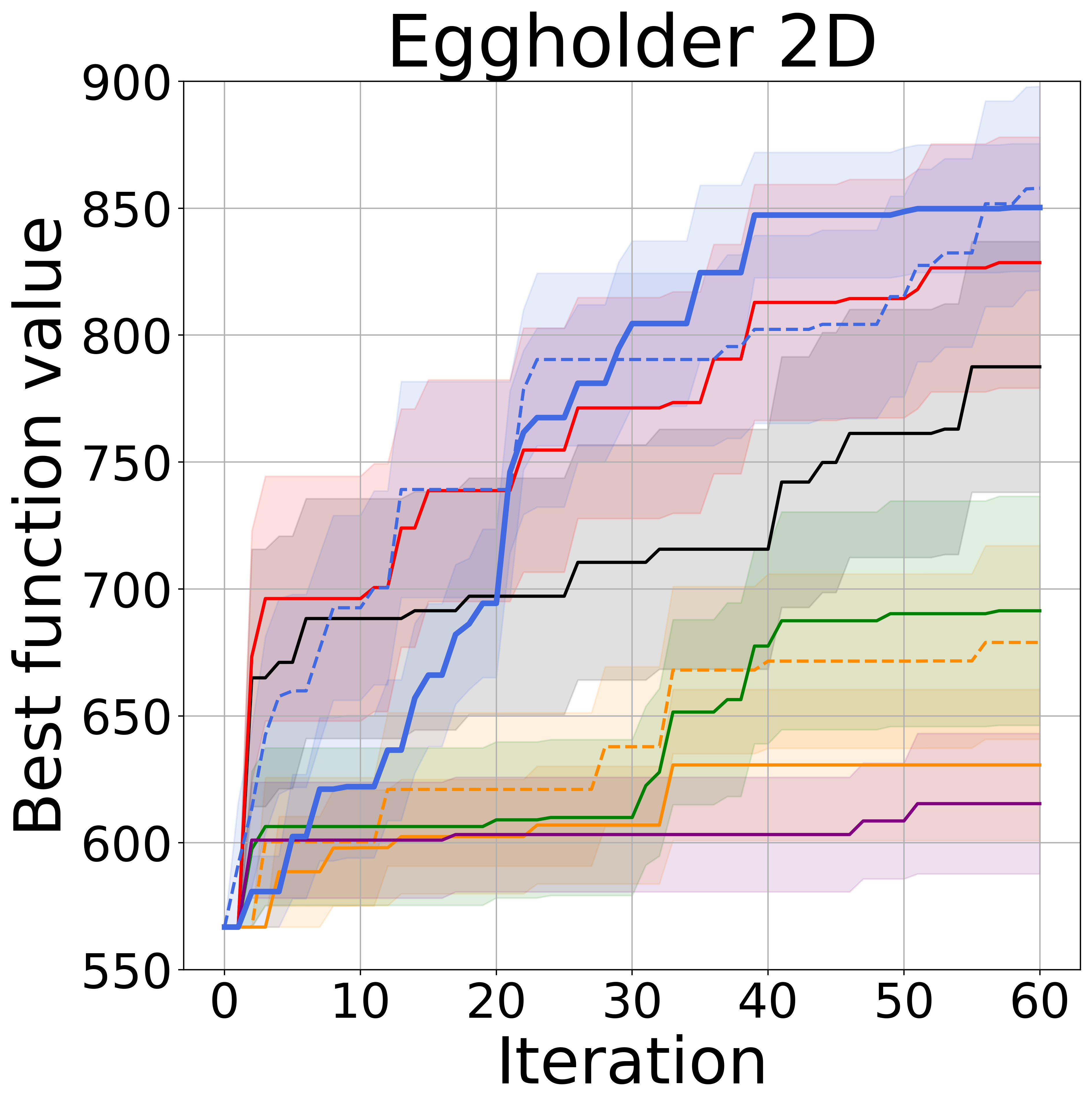}
\par\end{centering}
}\hspace{-0.2cm}\subfloat[\label{fig:performSynt_b}]{\begin{centering}
\includegraphics[scale=0.12]{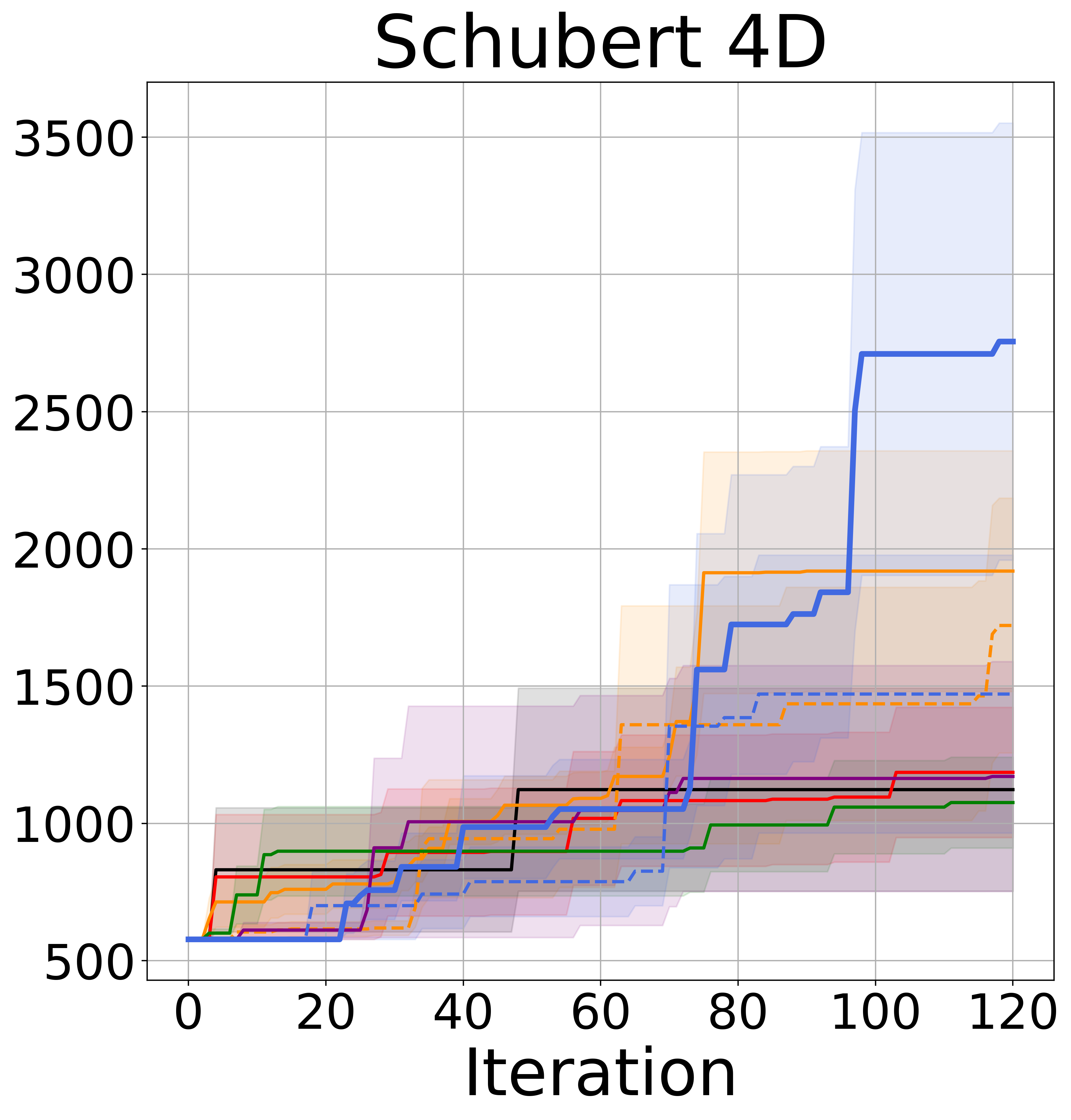}
\par\end{centering}

}\hspace{-0.2cm}\subfloat[\label{fig:performSynt_c}]{\begin{centering}
\includegraphics[scale=0.12]{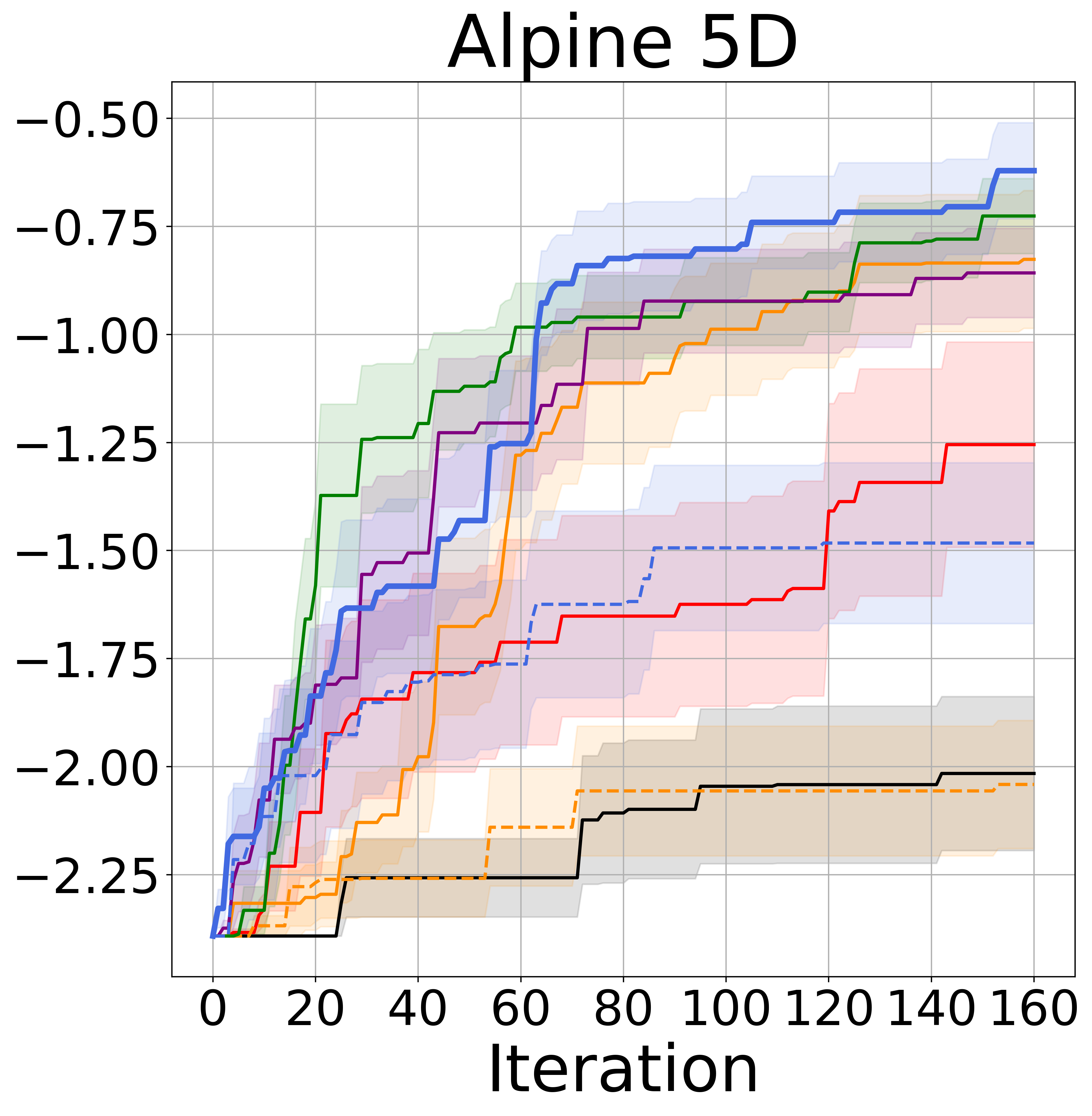}
\par\end{centering}

}\hspace{-0.2cm}\subfloat[\label{fig:performSynt_d}]{\begin{centering}
\includegraphics[scale=0.12]{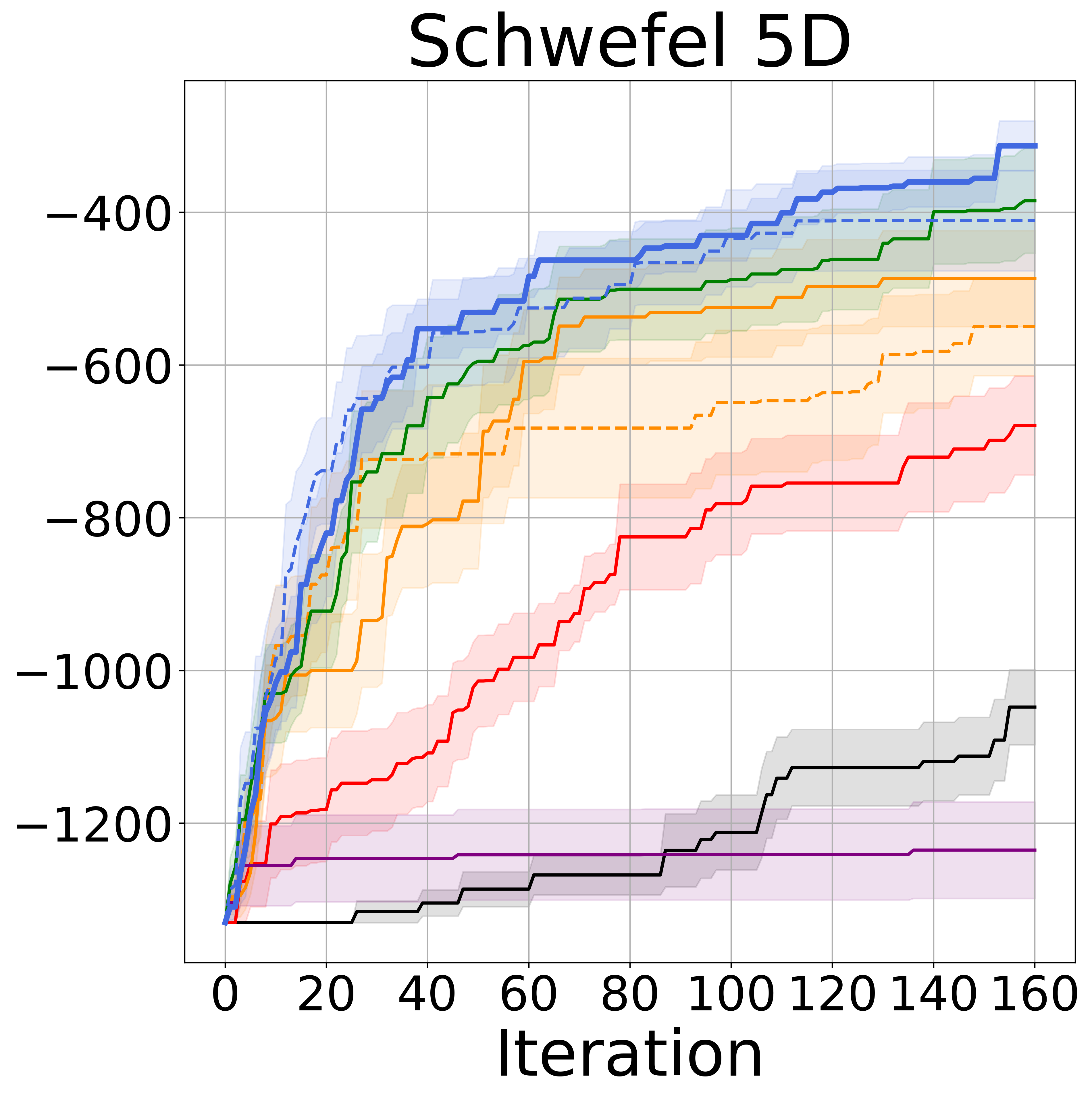}
\par\end{centering}
}\caption{\label{fig:performanceSynt}Optimization results for four synthetic
functions in Table \ref{table1:syntfuncchar}.}
\end{figure}

\textbf{Results and discussion.} Figure \ref{fig:performanceSynt}
shows the optimization results for four synthetic functions in Table
\ref{table1:syntfuncchar}. We can see that our method \textbf{BOMI}
generally outperforms other methods. On the 2d-function \textit{Eggholder}
(Figure \ref{fig:performanceSynt}(a)), \textbf{BOMI} and Imputation-BPMF
are the two best methods, where they are slightly better than SuggestBO.
When the dimension is increased up to 4 and 5 (Figure \ref{fig:performanceSynt}(b-d)),
\textbf{BOMI} is always the best method, and especially it significantly
outperforms other baselines on the 4d-function \textit{Schubert} (Figure
\ref{fig:performanceSynt}(b)). 

Imputation-based methods (mean, mode, KNN, and BPMF) work fairly well;
however, their performance is not very consistent. Mean and mode imputations
often fall behind KNN since they suffer from biases. Imputation-BPMF
is often better or comparable with other imputation methods, which
verifies our intuition about the importance of distributions of missing
values, as mentioned in Section \ref{subsec:Probability-missing}. \textbf{BOMI}
is often better than Imputation-BPMF. This clearly proves that our
proposal of using probability distributions to impute missing values
along with the new acquisition function is more effective, as discussed
in Section \ref{subsec:BOMI}.

DropBO underperforms on most functions since it throws away many observations
when they contain missing values, which leads to too few data to train
a good GP model. In contrast, SuggestBO is always better than DropBO
since it has more observations by replacing the missing value with
the value of suggested point. On the \textit{Eggholder} function (Figure
\ref{fig:performanceSynt}(a)), SuggestBO achieves a very good performance,
where it is the second-best method. However, on other functions SuggestBO
only achieves fair results since these functions vary very quickly
even with a small change in the input values. As expected, BO-uGP
unsuccessfully optimizes most functions due to its lack of the ability
to handle missing values.

\subsubsection{Stability comparison.}

The second experiment illustrates how different values of three factors
\textit{missing rate} $\rho$, \textit{missing noise} $\eta$, and
\textit{maximum number of missing values} $v$ affect to our method
and other baselines. Note that $\rho$ and $\eta$ were defined in
the first experiment setting, while $v$ indicates how many dimensions
in a data point contain missing values.

\textbf{Experiment settings.} We show the optimization result on the
5d-function \textit{Schwefel} as a function of one chosen factor while
the others are fixed to their default values. We sequentially set
up three separate settings as follows:
\begin{enumerate}
\item \textbf{Missing Rate.} We fix $\eta=0.05$ and $v=1$, then let $\rho\in[0.25,0.65]$
with a step of 0.1.
\item \textbf{Missing Noise.} We fix $\rho=0.25$ and $v=1$, then let $\eta\in[0.1,0.9]$
with a step of 0.1.
\item \textbf{Maximum number of missing values. }We fix $\rho=0.5$ and
$\eta=0.05$, then allow $v$ in a range of $[1,d-1]$, where $d=5$
is the dimension of function.
\end{enumerate}
\begin{figure}
\begin{centering}
\includegraphics[scale=0.24]{imgs/legendAnalysis_v2}\\
\subfloat[\label{fig:avg_per_synt_a}]{\begin{centering}
\includegraphics[scale=0.155]{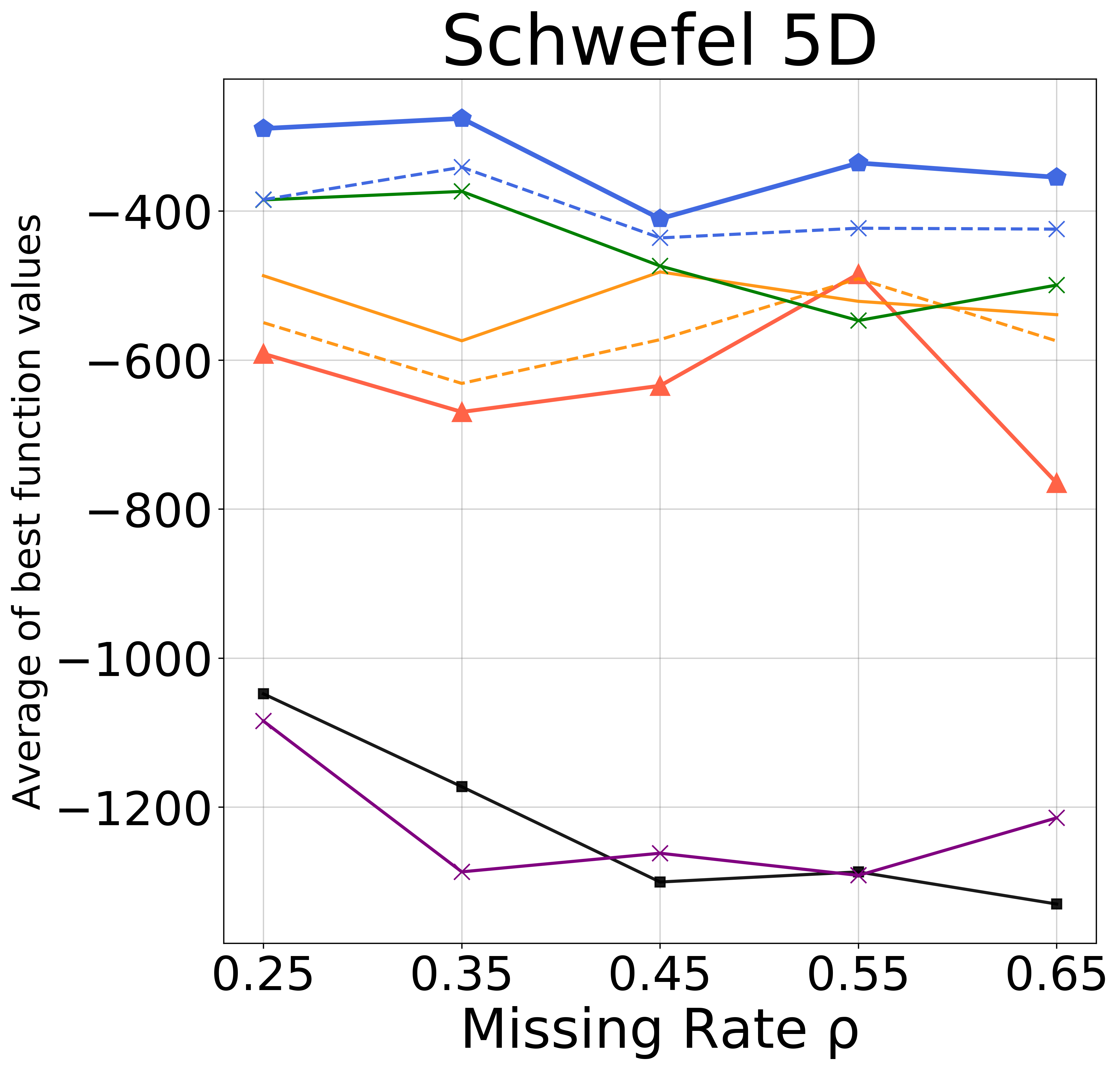}
\par\end{centering}
}\subfloat[\label{fig:avg_per_synt_b}]{\begin{centering}
\includegraphics[scale=0.155]{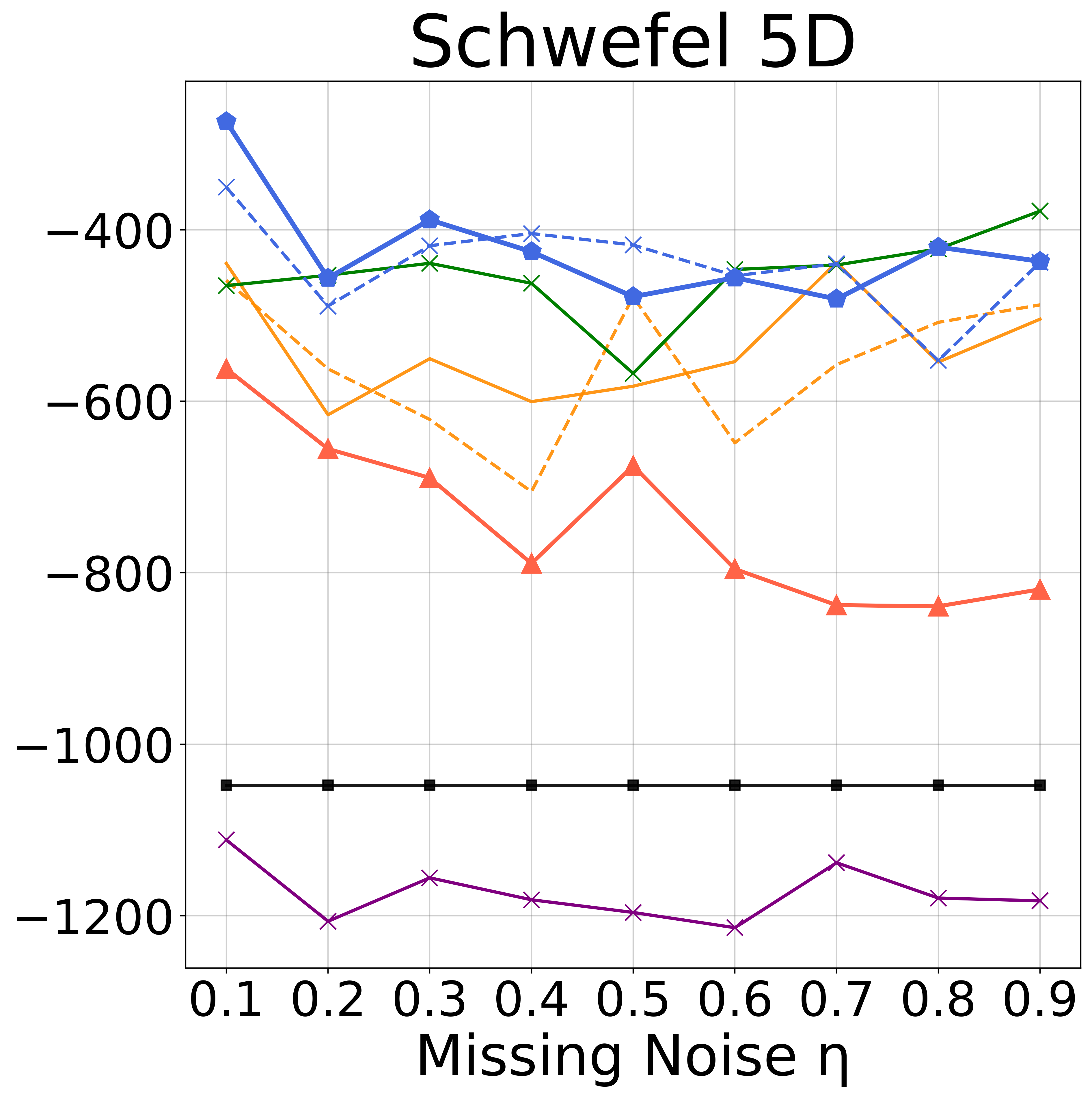}
\par\end{centering}

}\subfloat[\label{fig:avg_per_synt_c}]{\begin{centering}
\includegraphics[scale=0.155]{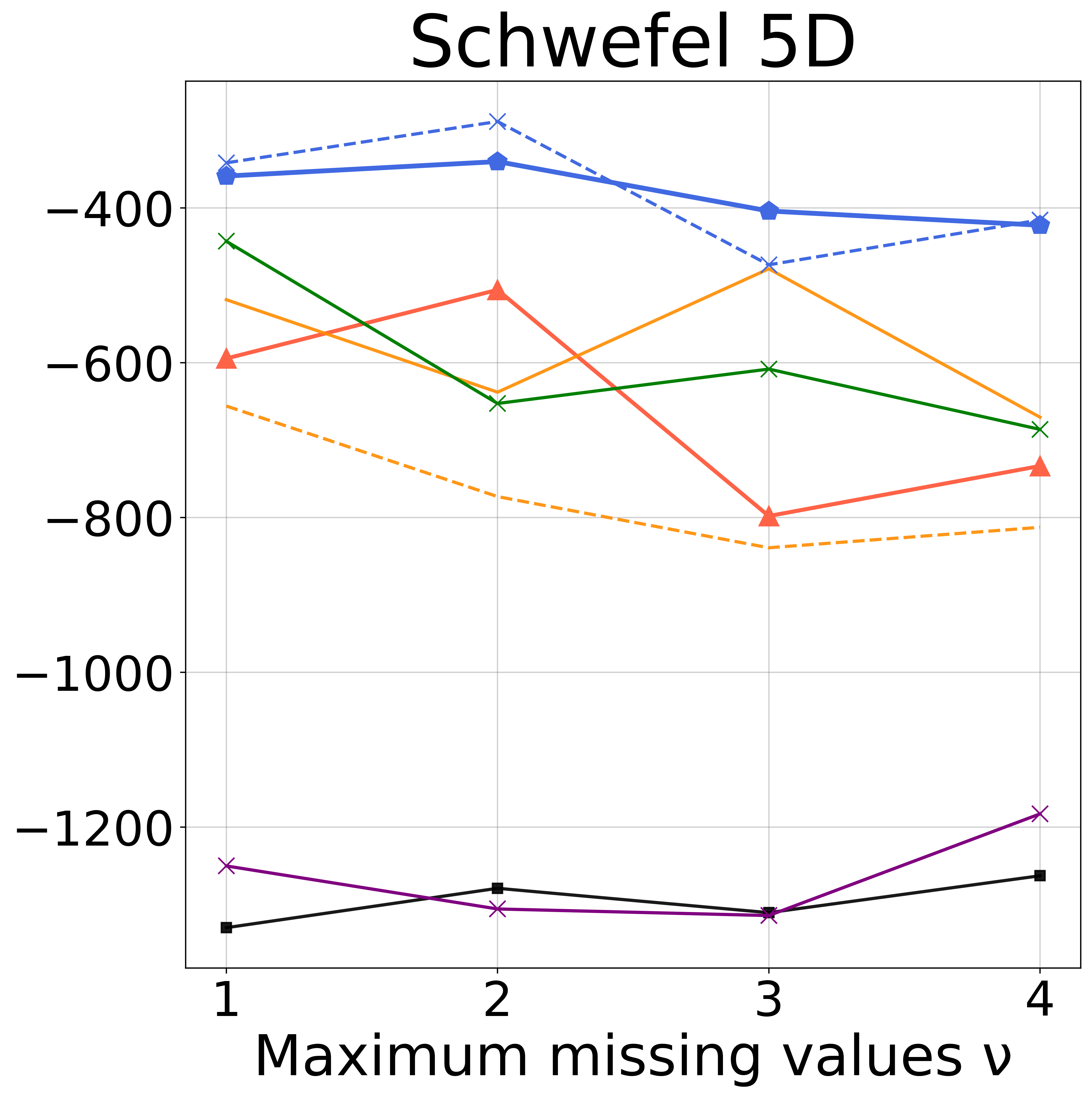}
\par\end{centering}
}
\par\end{centering}
\caption{\label{fig:result-test-stability}Optimization results of our method
\textbf{BOMI} and other methods on the 5d-fucntion \textit{Schwefel}
with different values for (a) missing rate $\rho$, (b) missing noise
$\eta$, and (c) maximum number of missing values $v$.}
\end{figure}

\textbf{Results and discussion.} From Figure \ref{fig:result-test-stability}(a),
we can see that when $\rho$ increases, the performance of DropBO
drastically declines. This can be explained by the fact that the number
of observations in DropBO is inversely proportional to $\rho$, which
leads to too few data to train a good optimization model. Similarly,
BO-uGP also faces the same problem as DropBO since it has no mechanism
to handle missing values. Meanwhile, SuggestBO seems to be unstable,
where its performance drops at $\rho=0.35$ and $0.45$ but increases
at $\rho=0.55$ before going down again at $\rho=0.65$. In contrast,
imputation-based methods and our method \textbf{BOMI} are stable and
robust to the missing rate, where the performance is just slightly
changed with different values for $\rho$.

From Figure \ref{fig:result-test-stability}(b), when $\eta$ increases
SuggestBO heavily drops since it imputation error increases in proportional
to $\eta$. Interestingly, the performance of DropBO does not change
since the noise is only applied to observations with missing values
and DropBO does not consider these observations. BO-uGP is the worst
method in this experiment since it computes wrong probability distributions
of very noisy values. Imputation-based methods except KNN wiggles
a lot, indicating that they suffer from an over-fitting. In contrast,
our method\textbf{ BOMI} can maintain a good performance even with
very high values for missing noise (e.g. $\eta=0.8,0.9$).

Finally, when the number of missing values increases (Figure \ref{fig:result-test-stability}(c)),
all methods trend to decrease, as expected. When more values are missing,
the correlation vanishes that, in turn, reduces the optimization performance.
Our method is still the best method, where it significantly outperforms
other methods. Similar to Figures \ref{fig:result-test-stability}(a-b),
DropBO and BO-uGP perform poorly in this experiment, where they are
the two worst methods.

\subsection{Real-world experiments}

We also demonstrate the benefits of our method in two real-world applications,
namely, robot exploration simulation \cite{rohmer2013v} and heat
treatment process \cite{gupta2018exploiting}.

\subsubsection{Robot exploration simulation.}

We use the simulation software named CoppeliaSim\footnote{\url{https://www.coppeliarobotics.com/}}
v.4 to simulate an environment for a robot to explore and measure
the concentration of copper in the soil \cite{rohmer2013v}. The environment
is created by using the dataset Brenda Mines\footnote{\url{http://www.kriging.com/datasets/}},
which includes a textured terrain, trees, and bumps. \textit{Our goal
is to find the best configuration for the robot to obtain the highest
percentage of copper.}

\begin{figure}
\begin{centering}
\subfloat[\label{fig:robotsim_a}]{\centering{}\includegraphics[scale=0.38]{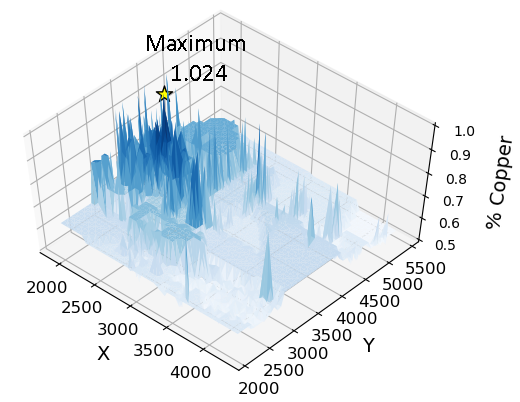}}\hspace{0.4cm}\subfloat[\label{fig:robotsim_b}]{\centering{}\includegraphics[scale=0.25]{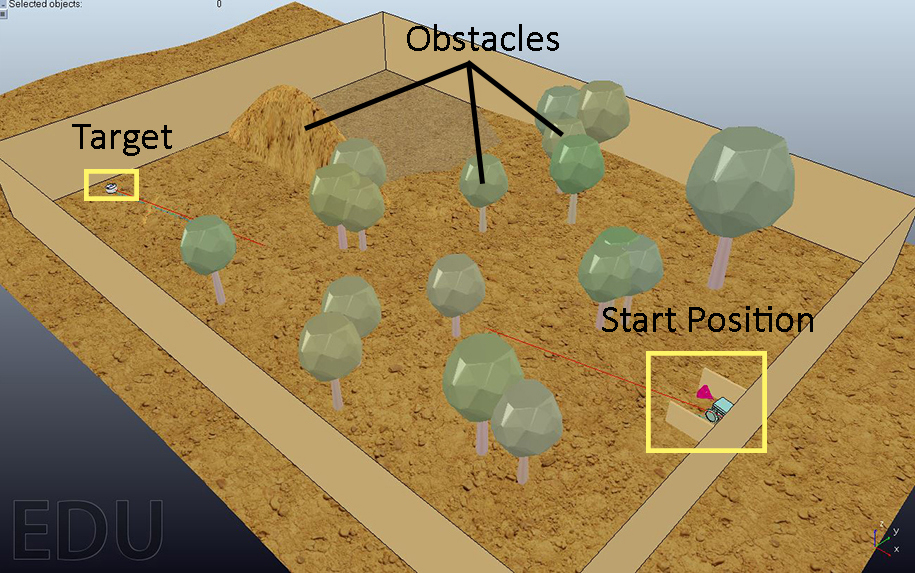}}
\par\end{centering}
\caption{\label{fig:coppeliaScreenShot}(a) A visualization of copper density
in the dataset Brenda Mines -- the darker blue indicates a location
with more copper, and the highest copper percentage is 1.024. (b)
A screenshot of the robot simulation using CoppeliaSim v.4 software
\cite{rohmer2013v} -- the figure shows the starting location of
the robot, its target location (i.e. \textit{the next location it
needs to move to}), and obstacles in the environment, e.g. trees and
bumps.}
\end{figure}

Figure \ref{fig:coppeliaScreenShot}(a) visualizes the copper percentage
in the dataset Brenda Mines, where the highest copper percentage is
1.024. This map is matched with the area shown in Figure \ref{fig:coppeliaScreenShot}(b),
where the robot needs to explore. Figure \ref{fig:coppeliaScreenShot}(b)
shows the starting location of the robot, its target location (i.e.
its \textit{next location}), and also obstacles (e.g. trees and bumps).
A 2-wheel robot is allowed 10 seconds to move along a pre-calculated
path to a specific next location. If the robot is unable to reach
the target location within 10 seconds, it takes the measurement at
the current location before the simulation stops. On the way to the
target location, many errors such as overturn or being stuck can happen
and prevent the robot from reaching the target location. Whenever
these errors occur, a certain noise is added to the current location
of the robot by the simulation software.

In this experiment, we tune four parameters $X$, $Y$, $Z$, and
$velocity$ of the robot. $X\in\left[1899.94,4301.51\right]$ and
$Y\in\left[2177.37,5400.19\right]$ are the coordinators of the next
location where the robot needs to move to. $Z\in\text{\ensuremath{\left[4330.96,5467.46\right]}}$
is the depth underground that the robot needs to drill to measure
the copper percentage at the location $(X,Y)$. $velocity\in\left[200,700\right]$
is the speed of robot moving; the value range of velocity is chosen
according to the simulation and path finding algorithm. We use the
missing rate $\rho=0.5$ and the number of missing variables $v=1$
(i.e. one of two coordinators of the robot can be missing). We do
not set value for the missing noise $\eta$ since the noise is automatically
added by the simulation when errors happen.

\begin{figure}[h]
\begin{centering}
\includegraphics[scale=0.24]{imgs/legendAnalysis_v2}\\
\subfloat[\label{fig:real_result_a}]{\begin{centering}
\includegraphics[scale=0.245]{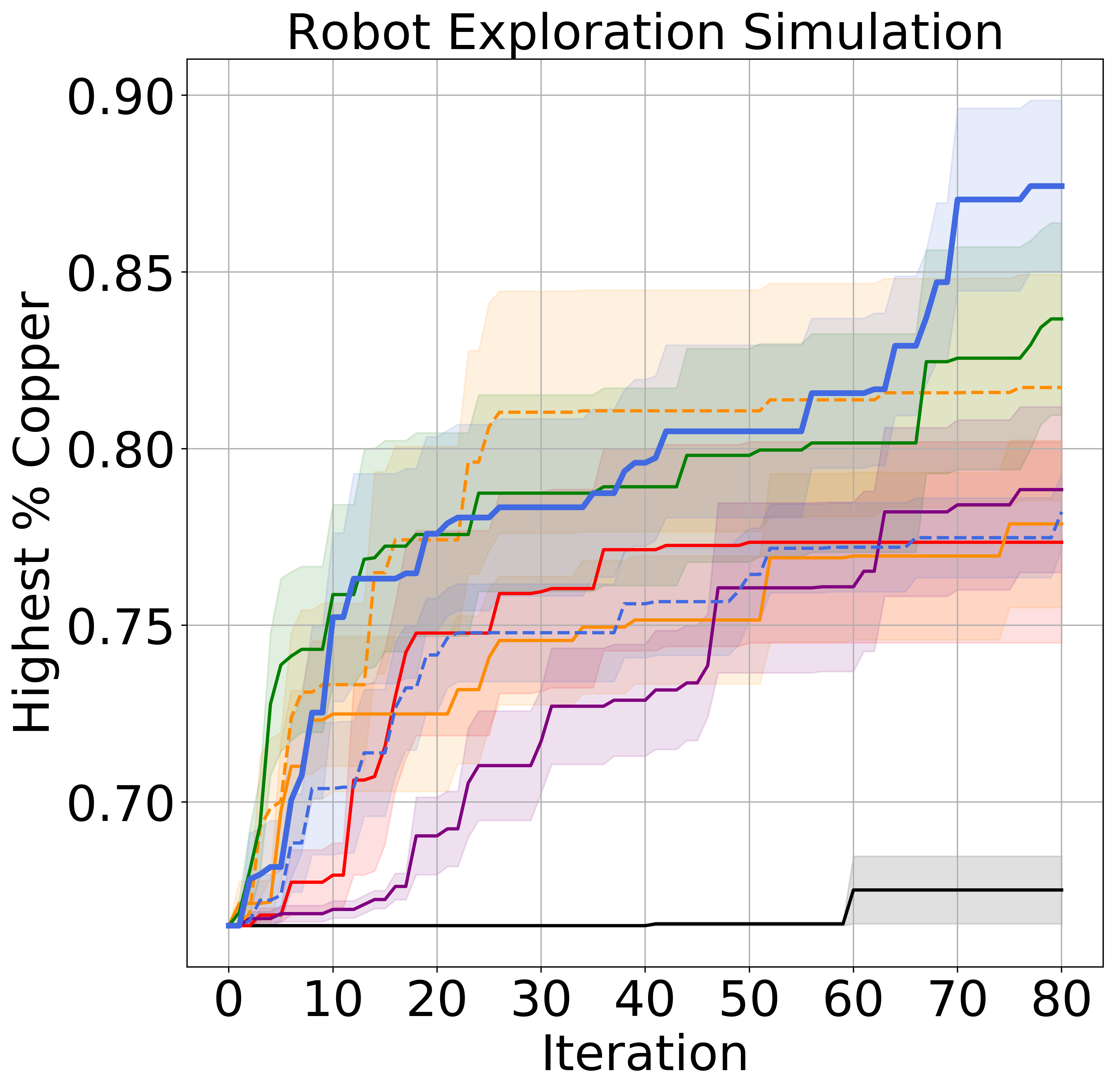}
\par\end{centering}
}\subfloat[\label{fig:real_result_b}]{\begin{centering}
\includegraphics[scale=0.245]{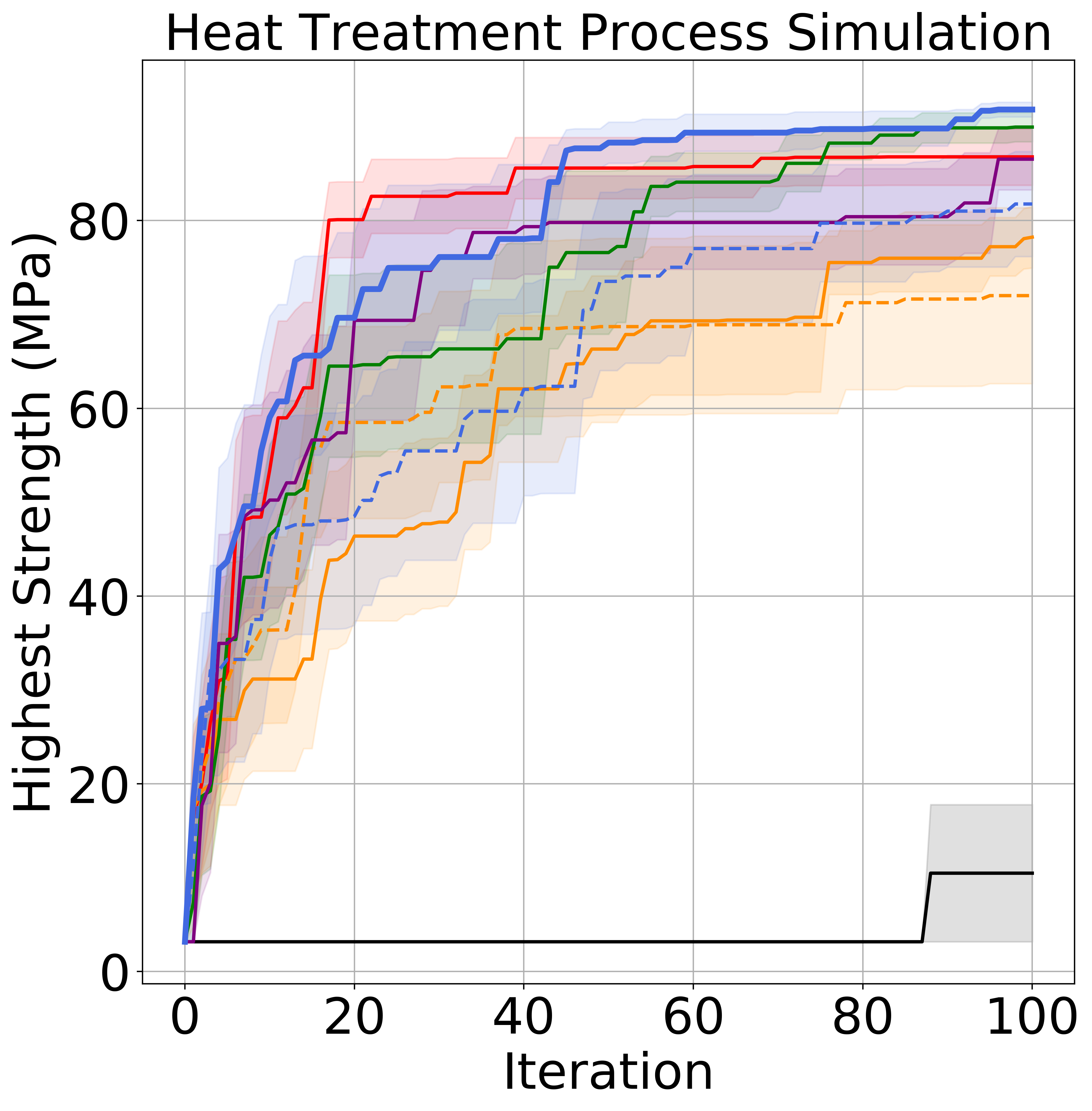}
\par\end{centering}
}
\par\end{centering}
\caption{\label{fig:results-on-real-world}Optimization results on two real-world
applications: (a) robot exploration simulation and (b) heat treatment
process. }
\end{figure}

From the result in Figure \ref{fig:results-on-real-world}(a), we
can see our method \textbf{BOMI} performs the best, where it significantly
outperforms other methods after 70 iterations. Two imputation-based
methods Imputation-KNN and Imputation-Mode perform well in this experiment,
where Imputation-KNN is the second-best method. Interesting, BO-uGP
shows a good performance in this application, where it is better than
SuggestBO and two other imputation-based methods. Again, the performance
of DropBO is very poor.

\subsubsection{Heat treatment process.}

This is a process of heating an alloy to achieve a desired strength.
In particular, an Al-Sc alloy is posed to the heat in four stages,
where each stage has a different temperature and a different time
duration. \textit{Our goal is to choose which temperature and how
long to heat the alloy at each stage to maximize its strength.} 

To simulate the heat treatment process for Al-Sc alloy, we use the
Kampmann-Wagner model \cite{wagner1991homogeneous}, same as in \cite{gupta2018exploiting}.
At each stage, there are two values to set, temperature $te$ (in
$\lyxmathsym{\textdegree}C$) and time $ti$ (in second). In total,
we tune eight parameters, including $te_{1}\in\left[1,100\right],te_{2}\in\left[1,1000\right],te_{3}\in\left[1,1000\right],te_{4}\in\left[1,1000\right]$,
and the heating times $ti_{k}\in\left[1,21600\right]$ for $k\in\left\{ 1,2,3,4\right\} $.
Since both temperature and heating time can be missing, we set the
missing rate and missing noise $\rho_{1}=0.35$, $\eta_{1}=0.8$ for
temperature and $\rho_{2}=0.25$, $\eta_{2}\in\left[0.7,0.9\right]$
for heating time.

Figure \ref{fig:results-on-real-world}(b) shows the optimization
result for the heat treatment process. We can see our method \textbf{BOMI}
is the best method, where it slightly outperforms the second-best
method Imputation-KNN. Two BO-based methods SuggestBO and BO-uGP perform
well and become the third-best method. It can be seen that BO methods
are generally better than imputation-based methods in this experiment.
We also see DropBO performs very poorly; it can be concluded that
this method is not favorable in practice.

The results in Figure \ref{fig:results-on-real-world} again confirm
the real benefits of our method not only in synthetic applications
but also in real-world applications when optimizing black-box functions
with missing inputs.

\section{Conclusion}

\label{sec:conclusion}We have presented a novel BO method \textbf{BOMI} to optimize expensive
black-box functions with missing values in inputs. Our method computes
the distributions of missing values for imputation and develops a
new acquisition function that takes into account the uncertainty of
imputed values to suggest the next point with more confidence. We
demonstrate the efficiency of \textbf{BOMI} with several benchmark
synthetic functions and two real-world applications in robot exploration
simulation and heat treatment process. The empirical results show
that \textbf{BOMI} has a better and more stable performance compared
to state-of-the-art baselines, especially in experiments with high
missing rates. Our future work will focus on improving the prediction
of missing values, which can help to improve the performance of our
method.

\section*{Acknowledgements}

This research was partially funded by the Australian Government through
the Australian Research Council (ARC). Prof Venkatesh is the recipient
of an ARC Australian Laureate Fellowship (FL170100006).


\end{document}